\documentclass{article}

\usepackage{arxiv}
\usepackage[utf8]{inputenc} 
\usepackage[T1]{fontenc}    
\usepackage{hyperref}       
\usepackage{url}            
\usepackage{booktabs}       
\usepackage{amsfonts}       
\usepackage{nicefrac}       
\usepackage{microtype}      
\usepackage{lipsum}
\usepackage{graphicx}
\usepackage{wrapfig}
\usepackage{array}
\usepackage[export]{adjustbox}
\usepackage{amsmath}
\usepackage[numbers,sort&compress]{natbib}
\graphicspath{ {./images/} }

\title{LTV-YOLO: A Lightweight Thermal Object Detector for Young Pedestrians in Adverse Conditions}

\author{
 Abdullah Jirjees \\
  Automotive and Surface Transportation\\
  National Research Council Canada\\
  London, ON, Canada\\
  \texttt{abdullah.jirjees@nrc-cnrc.gc.ca} \\
   \And
 Ryan Myers \\
  Automotive and Surface Transportation\\
  National Research Council Canada\\
  London, ON, Canada\\
  \texttt{ryan.myers@nrc-cnrc.gc.ca} \\
  \And
 Muhammad Haris Ikram \\
  Department of Civil and Environmental Engineering\\
  University of Western Ontario\\
  London, ON, Canada\\
  \texttt{mikram5@uwo.ca} \\
  \And
 Mohamed H. Zaki \\
  Department of Civil and Environmental Engineering\\
  University of Western Ontario\\
  London, ON, Canada\\
  \texttt{m.zaki@uwo.ca} \\
}

\begin{document}
\maketitle
\begin{abstract}
Detecting vulnerable road users (VRUs), particularly children and adolescents, in low light and adverse weather conditions remains a critical challenge in computer vision, surveillance, and autonomous vehicle systems. This paper presents a purpose-built lightweight object detection model designed to identify young pedestrians in various environmental scenarios. To address these challenges, our approach leverages thermal imaging from long-wave infrared (LWIR) cameras, which enhances detection reliability in conditions where traditional RGB cameras operating in the visible spectrum fail. Based on the YOLO11 architecture and customized for thermal detection, our model, termed LTV-YOLO (Lightweight Thermal Vision YOLO), is optimized for computational efficiency, accuracy and real-time performance on edge devices. By integrating separable convolutions in depth and a feature pyramid network (FPN), LTV-YOLO achieves strong performance in detecting small-scale, partially occluded, and thermally distinct VRUs while maintaining a compact architecture. This work contributes a practical and scalable solution to improve pedestrian safety in intelligent transportation systems, particularly in school zones, autonomous navigation, and smart city infrastructure. 
Unlike prior thermal detectors, our contribution is \emph{task-specific}: a thermally only edge-capable design designed for young and small VRUs (children and distant adults). Although FPN and depthwise separable convolutions are standard components, their integration into a thermal-only pipeline optimized for short/occluded VRUs under adverse conditions is, to the best of our knowledge, novel.
\end{abstract}

\keywords{Thermal Imaging, YOLO, Lightweight Object Detection, Vulnerable Road Users, Edge Computing, Autonomous Vehicles}

\section{Introduction}
The global burden of road traffic deaths remains a critical concern for transportation safety. According to the World Health Organization (WHO), more than 21\% of traffic-related deaths involve pedestrians, with children and adolescents among the most vulnerable road users (VRUs) \cite{world2019global}. Young VRUs face disproportionately higher risk due to their smaller physical profiles, unpredictable behavior, and limited visibility in challenging environments such as night, rain, snow, or fog. These challenges are exacerbated by current pedestrian detection systems that rely heavily on RGB imaging, which suffers in low light and adverse weather \cite{guan2019fusion,lee2025multispectral}.

Recent analyses show that low-light conditions are among the most dangerous times for VRUs. In Canada, more than 1 in 4 pedestrian fatalities (26\%) occur at night, while weather, poor road conditions, and infrastructure issues contribute to 23\% of fatal incidents \cite{statcan2023pedestrian}. Furthermore, 9\% of pedestrian deaths involved individuals wearing dark clothing, underscoring the need for detectors that operate reliably in low-visibility real-world environments.

Deep learning–based object detectors such as YOLO \cite{hidayatullah2025yolov8,tian2025yolov12,yolo11_ultralytics}, SSD \cite{liu2016ssd}, and Faster R-CNN \cite{ren2016faster} have improved general pedestrian detection. However, their effectiveness declines on edge devices in real driving conditions, especially for detecting young/short pedestrians \cite{xu2019learning,7298706}. Thermal imaging using long-wave infrared (LWIR) sensors offers a promising alternative: by measuring heat rather than reflected visible light, thermal cameras provide strong human–background contrast at night and in poor weather \cite{zhang2019cross}.

Thermal imaging also presents challenges. Thermal frames lack the color/texture cues leveraged by CNNs trained on RGB data, and available thermal datasets are smaller and less diverse, increasing overfitting risk. Many thermal approaches, therefore, adopt heavier architectures that are impractical for real-time inference on resource-constrained edge devices.

To address these limitations, we propose LTV-YOLO, a lightweight thermal-vision detector designed to enhance the detection of young VRUs in low-visibility environments, combining the robustness of thermal imaging with a compact architecture optimized for edge deployment. Our key contributions are: (1) a purpose-built lightweight thermal detector for young/short VRUs, (2) an edge-optimized architecture with real-time performance, and (3) experimental validation in adverse environmental conditions.

\noindent\textbf{Novelty in task-specific integration.}
We do not claim new layers; instead, our novelty lies in \emph{how} known components are assembled for this problem: (i) a privacy-preserving, \emph{thermal-only} pipeline; (ii) small-object and occlusion-aware design focused on \emph{young/short VRUs}; and (iii) aggressive edge optimization (latency, memory) for real-time operation on Jetson-class devices.

\section{Related Work}
\subsection{Object Detection Models for VRU Detection}
Deep learning-based object detection models have revolutionized computer vision, enabling significant advances in pedestrian detection. Models such as YOLO \cite{hidayatullah2025yolov8}, SSD \cite{liu2016ssd}, and Faster R-CNN \cite{ren2016faster} have been widely used for real-time detection of road users. Although these models offer competitive performance in standard datasets such as COCO \cite{lin2014microsoft} and CityScapes \cite{cordts2015cityscapes}, their generalization to vulnerable road users (VRUs), particularly children and adolescents, remains limited. Traditional datasets lack age-specific annotations and often fail to represent young pedestrians across various environmental conditions. As a result, these models may struggle to accurately detect smaller or partially occluded subjects, especially in high-risk areas, such as school zones and residential streets.

Partial occlusion remains a critical failure mode in real-world pedestrian detection, where children may be obscured by vehicles, poles, or other pedestrians. Several studies have addressed occlusion-aware detection using multiscale features, context reasoning, or attention mechanisms \cite{zhang2020feature,wang2018pcn}.

\subsection{Thermal-Based Object Detection}
Thermal imaging has emerged as a robust sensing modality for detecting vulnerable road users (VRUs) in low-light and adverse weather conditions. Unlike RGB cameras, thermal sensors capture long-wave infrared (LWIR) radiation, producing consistent human silhouettes regardless of ambient lighting \cite{zhang2019cross}. Publicly available datasets have catalyzed research into fusion-based and standalone thermal pedestrian detection systems.

A significant body of work has investigated RGB–thermal fusion techniques to improve detection accuracy across diverse conditions \cite{guan2019fusion}. In parallel, other studies have focused exclusively on thermal inputs to preserve privacy and enhance robustness under poor illumination \cite{xu2019learning}. However, many of these approaches rely on computationally intensive backbones such as ResNet or EfficientNet, limiting their practicality for real-time applications on embedded or resource-constrained edge devices \cite{Howard_2019_ICCV,tan2019efficientnet,li2023tfdet}.

Furthermore, while recent efforts such as RGBT-Tiny \cite{ying2024visible} and TFDet \cite{li2023tfdet} explore thermal detection for small objects or in adverse weather conditions, few models are explicitly designed to address the unique detection challenges posed by ``young VRUs.'' These include smaller physical size, lower thermal contrast, and dynamic behavior—all of which degrade detection performance in typical pedestrian detectors. This highlights a critical research gap in the development of lightweight, thermal-only object detectors capable of operating under real-world conditions and optimized for detecting children in safety-critical environments.

\subsection{Lightweight Object Detection Models}
To address the computational limitations of the deployment in embedded systems, lightweight models such as YOLOv5n \cite{jocher2020ultralytics}, YOLOv8n \cite{hussain2023yolo}, MobileNet-SSD \cite{howard2017mobilenets}, and Tiny-YOLO \cite{9128749} variants have gained traction. These models reduce parameter count and memory usage by using techniques such as depth-wise separable convolutions and simplified detection heads. However, while these models are designed for speed and efficiency, they often sacrifice accuracy when applied to difficult scenarios such as thermal imagery or small object detection \cite{tumas2021improvement,mehta2021mobilevit}. Few existing lightweight detectors are tailored to thermal environments or evaluated against young pedestrian benchmarks, highlighting a critical gap in edge-capable safety systems \cite{10382506}.

\subsection{Open Research Challenges}
Despite progress, several open challenges remain. First, existing pedestrian datasets under-represent children and adolescents, particularly in thermal and nighttime scenes. Second, robust pedestrian detection under diverse environmental conditions remains an unsolved problem for many visual systems. Third, integrating high-accuracy models with low-power real-time inference platforms is essential for wide-scale deployment in intelligent transportation systems. These gaps motivate the development of purpose-built solutions, such as LTV-YOLO, designed to detect young VRUs using thermal input while remaining efficient enough for edge-based inference on platforms like the Jetson Orin AGX.

\subsection{State-of-the-Art Thermal and RGB-Thermal Fusion}
Thermal pedestrian detection has advanced through thermal-only models such as TFDet\cite{li2023tfdet} and RGB–thermal fusion methods such as RGBT-Tiny\cite{ying2024visible}. Public datasets, including KAIST Multispectral\cite{hwang2015multispectral} and FLIR ADAS~\cite{flir2021dataset}, provide benchmarks for evaluation across diverse conditions. 

While these approaches improve general pedestrian detection under low-light and adverse weather, they typically employ heavier backbones and focus on broader adult pedestrian benchmarks. In contrast, our work focuses on a compact, \emph{thermal-only} detector explicitly optimized for \emph{young/short VRUs}, with edge-capable efficiency suitable for real-time deployment on embedded platforms.

\section{Methodology}
\subsection{LTV-YOLO Model Architecture}
The proposed model, LTV-YOLO (Lightweight Thermal Vision YOLO), is a purpose-built object detector designed to run efficiently on edge platforms such as the NVIDIA Jetson Orin AGX, while maintaining high detection accuracy under adverse environmental conditions. Inspired by the YOLO (You Only Look Once) architecture, LTV-YOLO follows a single-shot detection paradigm, making predictions in a single forward pass to ensure low latency. Unlike large-scale detectors designed for datacenter deployment, LTV-YOLO is tailored for the constraints of real-world embedded systems, balancing accuracy, latency, and thermal-specific input.

The model consists of three main components:
\begin{itemize}
    \item A lightweight backbone using depthwise separable convolution blocks for efficient feature extraction.
    \item A feature pyramid network (FPN) for multi-scale feature fusion, improving performance on small-sized targets such as children.
    \item A detection head that produces bounding boxes and class scores using thermal image input.
\end{itemize}

\subsubsection{Depthwise Separable Convolutions}

To enable efficient real-time inference on edge platforms, LTV-YOLO replaces conventional layers with \textit{depthwise separable convolutions}. This technique factorizes a standard convolution into two separate operations: (1) a \textit{depthwise convolution}, which applies a single filter to each input channel independently, and (2) a \textit{pointwise convolution}, which uses $1\times1$ kernels to combine the outputs across channels.

This decomposition significantly reduces the number of learnable parameters and floating-point operations (FLOPs). Compared to standard convolution, depthwise separable convolutions require approximately 80–90\% fewer computations \cite{howard2017mobilenets}. This efficiency allows LTV-YOLO to maintain strong representational capacity while achieving low-latency inference suitable for edge deployment.

\subsubsection{Feature Pyramid Network for Small Objects}
Detecting young VRUs, especially children, presents unique challenges due to their relatively small physical size in the image frame, especially in wide-angle thermal video feeds. To address this, LTV-YOLO incorporates a lightweight \textit{Feature Pyramid Network (FPN)} \cite{8099589}, which improves detection of small-scale objects by merging semantic features across multiple resolutions.

\begin{figure*}[!t]
    \centering
    \includegraphics[width=\linewidth]{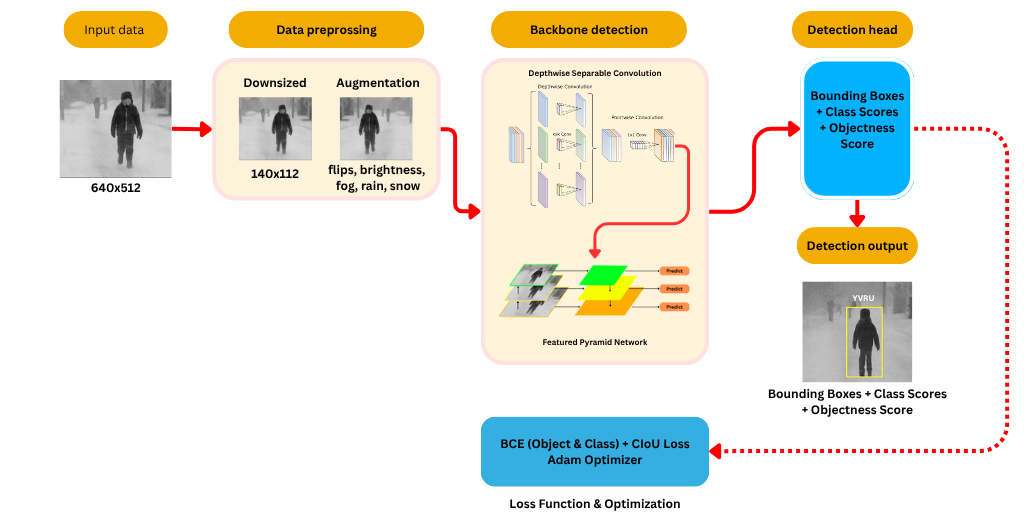}
    \caption{Thermal detection pipeline of LTV-YOLO. The system receives input from a thermal camera, processes it through lightweight convolution and depthwise blocks, merges multi-scale features via FPN, and outputs bounding boxes or alerts.}
    \label{fig:fpn_diagram}
\end{figure*}

The FPN fuses information from deeper, coarser layers with spatially finer shallow layers via upsampling and lateral connections. This structure enhances the model’s ability to detect small pedestrians in occlusion, at a distance, or in low-resolution thermal contexts. As shown in Figure~\ref{fig:fpn_diagram}, the hierarchical features are aggregated and fed into the detection head.

\subsubsection{Loss Function and Optimization}

LTV-YOLO is trained using a composite loss function designed to jointly optimize object classification and bounding-box localization. Specifically, we combine binary cross-entropy (BCE) loss for classification and Complete Intersection over Union (CIoU) \cite{zheng2020distance} loss for bounding box regression. This dual-objective approach allows the model to balance classification confidence with precise spatial alignment.

We optimize a composite loss:

\begin{equation}
\mathcal{L}_{\text{total}} = \lambda_{\text{obj}} \mathcal{L}_{\text{obj}} + \lambda_{\text{cls}} \mathcal{L}_{\text{cls}} + \lambda_{\text{loc}} \mathcal{L}_{\text{CIoU}},
\end{equation}

We used the following weighting: $\lambda_{\text{obj}} = 1$, $\lambda_{\text{cls}} = 1$, and $\lambda_{\text{loc}} = 5$.

The 5x emphasis on localization reflects the application’s safety-critical nature, where false negatives (missed detections) and poor bounding box quality can directly impact pedestrian safety. In ablation experiments, increasing the weight of CIoU loss improved bounding box alignment and reduced missed detections of small pedestrians.

CIoU is chosen over standard IoU or GIoU because it penalizes not only bounding box overlap but also the distance between centers and differences in aspect ratio—critical factors in pedestrian detection under varying poses and scales \cite{zheng2020distance}.

The network is optimized using the Adam optimizer with an initial learning rate of $10^{-3}$ and a cosine annealing schedule, which has been shown to improve convergence and generalization in deep learning models \cite{loshchilov2017sgdr}. Training is conducted for 100 epochs with a batch size of 32. 

To improve the generalization of the model to diverse real-world conditions, we apply targeted data augmentations. Random horizontal flipping simulates pedestrian appearances on both sides, for edge deployment perturbations introduce variation in thermal contrast, accounting for differences in ambient conditions. Synthetic fog and rain overlays mimic adverse weather scenarios, such as occlusion and visibility degradation—enhancing robustness under low-visibility conditions, where annotated data is scarce \cite{shorten2019survey,tremblay2018training}.

\subsection{Edge Deployment}

LTV-YOLO is designed with edge deployment in mind. The total model size is less than 10MB with approximately 4 million parameters. It runs at more than 50 frames per second on the Jetson AGX Orin 32GB, 2048-core GPU with thermal input from a FLIR ADK, 640 x 512, GMSL1, 24° FOV camera. The efficient computation pipeline and reduced memory footprint make it well-suited for real-time applications such as autonomous vehicles, smart surveillance, and school zone monitoring, where high-speed, low-power operation is essential.

To ensure high reliability in safety-critical scenarios, we set a minimum confidence threshold of 0.5 for triggering a detection alert. This threshold was calibrated empirically based on a precision-recall trade-off observed on the validation folds during 10-fold cross-validation. We selected the value that maximized the F1-score while maintaining a low false-negative rate, which is particularly important for detecting young pedestrians. In downstream applications such as ADAS or smart infrastructure, this threshold can be further tuned based on deployment-specific risk tolerances and response time requirements.

\subsubsection{Input Resolution and Preprocessing}

To balance detection accuracy with inference speed on edge devices, thermal input frames of resolution $640 \times 512$ are downsampled to $140 \times 112$ pixels prior to the inference phase. This resizing is performed using bilinear interpolation to preserve the structural integrity of small-scale pedestrian features while significantly reducing computational load. To optimize real-time performance on edge hardware, we empirically tested input resolutions of 224×179, 140×112, and 96×77, all of which preserve the original 1.25:1 aspect ratio of the thermal images (640×512). The 140×112 input provided the best trade-off between detection accuracy and inference speed on the Jetson Orin AGX, making it ideal for embedded deployment.

\section{Experimental Setup}
\subsection{Dataset}

To evaluate the performance of the proposed LTV-YOLO model, we constructed a custom thermal dataset consisting of approximately 10,000 images captured using a FLIR ADK camera (640×512 resolution, 24° FOV, GMSL1 interface). The dataset includes children and adult pedestrians to reflect general small VRUs encountered in residential and suburban areas. Recordings were conducted under diverse real-world conditions, including snowfall and daylight scenes.

Our proposed dataset was split using 10-fold cross-validation, maintaining a 50:50 balance between child and adult pedestrian instances in each fold. All annotations were manually labeled using the LabelMe annotation tool, including bounding boxes for visible VRUs. The categorization of pedestrians as children versus adults was performed subjectively based on visual appearance, such as height, body proportions, and clothing cues, in the absence of ground-truth age labels. This distinction was validated across multiple annotators to improve consistency.

\begin{figure}[!t]
\centering
\includegraphics[width=\linewidth]{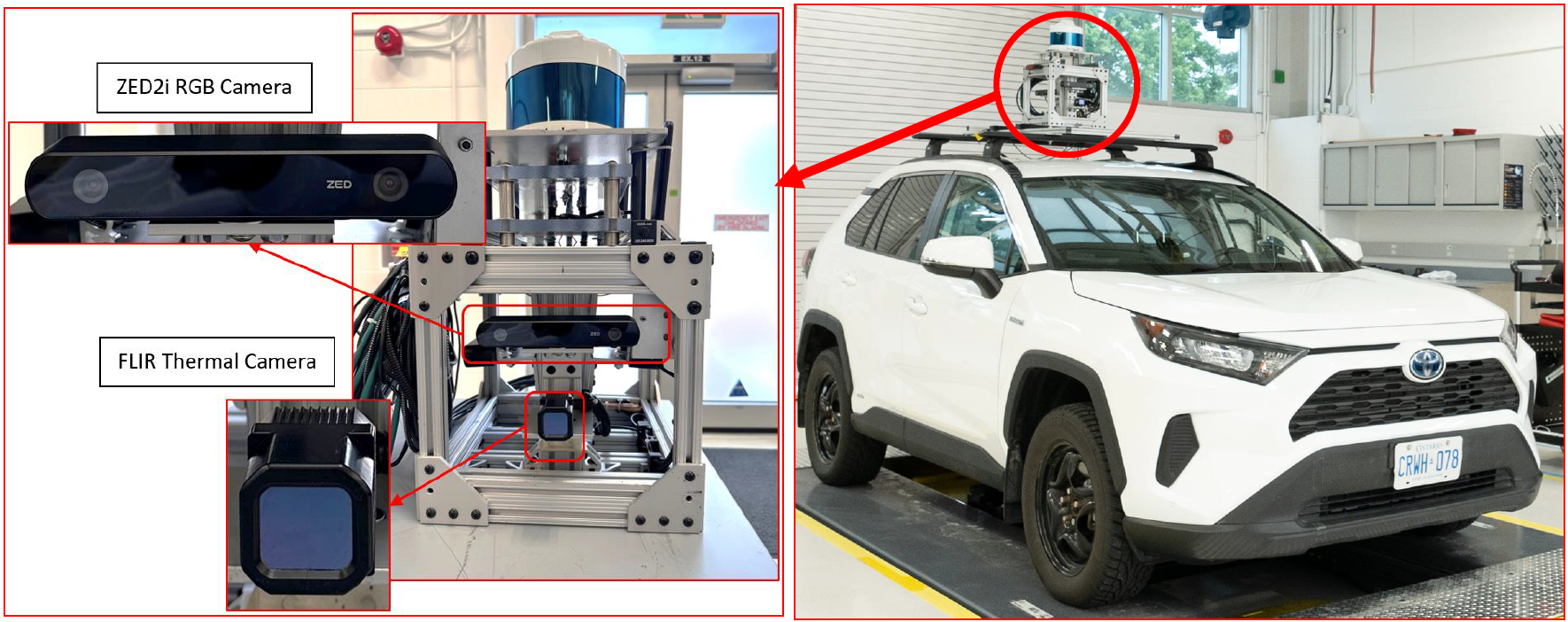}
\caption{Sensor module for thermal data collection: ZED2i (top) and FLIR ADK (bottom).}
\label{fig:sensor_rig}
\end{figure}

For synchronized multimodal data acquisition, a custom sensor rig was mounted on top of the research vehicle. The configuration included a ZED2i stereo RGB camera (top) and the FLIR ADK thermal camera (bottom), vertically aligned to ensure overlapping fields of view. Both sensors were synchronized to enable concurrent RGB and thermal recordings under identical conditions. Figure~\ref{fig:sensor_rig} illustrates the physical setup of the sensor module used during data collection.

In addition to our custom dataset, we also utilized the public thermal benchmarks LLVIP (Low Light Vision) \cite{jia2021llvip} and OpenThermalPose2 \cite{kuzdeuov2025openthermalpose2} to validate our model against existing models and ensure fair evaluation. LLVIP provides over 30,000 temporally and spatially aligned visible-infrared image pairs collected in low-light street scenes from surveillance cameras, serving as a challenging benchmark for nighttime pedestrian detection. OpenThermalPose2, on the other hand, is a large-scale human pose dataset containing over 11,391 thermal images across diverse activities and weather conditions. While these datasets provide a baseline for general human detection, they do not explicitly account for the unique scale and thermal profile of young VRUs, which distinguishes them from our purpose-built dataset.

\subsection{Training Details}
Model training was conducted using PyTorch on a laptop equipped with an NVIDIA RTX 4070 GPU. Inference benchmarking was performed on a Jetson AGX Orin (32GB, 2048-core GPU) to simulate deployment on embedded edge platforms. The training process followed the optimization and loss setup described in Section III (3). 

Training was conducted for 200 epochs with a batch size of 16. All models were trained with an input resolution of 640×512 to match the native resolution of the thermal camera. Weight decay was set to $5 \times 10^{-4}$, and training was performed using Batch Normalization (BN) layers.

To enhance the robustness of the model under diverse environmental conditions, extensive data augmentation was applied using the Albumentations library~\cite{info:albumentations}. The augmentation pipeline included random horizontal flipping with a probability of 0.5, brightness and contrast adjustments (0.3), and mosaic (1.0).

\subsection{Metrics}
We evaluated the performance of the model using standard object detection metrics. The mean Average Precision (mAP) was computed at an IoU threshold of 0.5 (mAP@0.5). Inference speed was measured in frames per second (FPS) on the Jetson AGX Orin using live thermal video inputs.

We also report the model’s total disk footprint and parameter count to assess its compactness. Energy efficiency was evaluated by monitoring power consumption during real-time inference using onboard Jetson profiling utilities.

To benchmark performance, the proposed LTV-YOLO model was compared with YOLOv5n, YOLOv8n, and YOLO11n under identical training configurations and data set splits. This facilitated a fair comparison of accuracy, inference speed, and resource efficiency under varying environmental conditions. We also compared against a small variant of the state-of-the-art transformer-based detector, the Real-Time Detection Transformer (RT-DETR) \cite{zhao2024detrs}. Due to the absence of the official implementation of the nano variant, we utilized the small variant as the closest comparable baseline.

\section{Experimental Evaluation}

\subsection{Quantitative Comparison}

To assess the effectiveness of the proposed LTV-YOLO model, we conducted a comparative analysis against YOLOv5n, YOLOv8n, YOLO11n and RT-DETRs, which are well-known lightweight object detectors. All models were trained and evaluated on the same thermal dataset, under identical experimental conditions.

We report performance using several key metrics: mean Average Precision at 0.5 IOU (mAP) on LLVIP, OpenThermalPose2, and a custom Young VRU subset; inference speed (frames per second, FPS) on the Jetson Orin; and model size.

\begin{table*}[t]
\centering
\caption{Performance on Jetson AGX Orin (32GB).}
\label{table:performance_comparison}
\resizebox{\textwidth}{!}{%
\begin{tabular}{|l|c|c|c|c|c|}
\hline
Model & Params (M) & mAP (LLVIP) & mAP (OpenThermalPose 2) & mAP (VRUs) & FPS \\
\hline
YOLOv5(n)     & 1.9  & 96.2\% & \textbf{99.3}\% & 84.5\% & 73 \\
YOLOv8(n)     & 3.2  & 96.5\% & \textbf{99.3}\% & 82.7\% & \textbf{77} \\
YOLO11(n)     & 2.6  & 96.6\% & 99.2\%          & 84.8\% & 67 \\
RT-DETR(s)    & 20   & \textbf{97.8\%} & 98.6\%   & 81.2\% & - \\
\textbf{LTV-YOLO} & \textbf{1.6} & \textbf{96.7}\% & 98.4\% & \textbf{91\%} & 75 \\
\hline
\end{tabular}%
}
\end{table*}

\subsection{Synthetic High-Temperature Backgrounds and Reflections}
Thermal false positives can arise from high-temperature backgrounds (e.g., sun-warmed asphalt or engine blocks) and from specular reflections on metallic surfaces. To simulate these effects, we augment training with: 
(i) \emph{temperature-bias} overlays (bright blobs near vehicles and road surfaces), 
(ii) \emph{specular strips} aligned with vehicle panels, and 
(iii) \emph{CutOut/CutMix} occluders applied to human torsos. 

Qualitatively, these augmentations reduced spurious bounding boxes on hot or reflective regions in validation images. Quantitatively, we observe a reduction in false positives per frame on a held-out “hot-bg” validation split, without degrading overall mAP. 

\subsection{Resolution Sensitivity}
To evaluate model resilience to input resolution, we tested LTV-YOLO on the same thermal video sequence at two different resolutions: the original $640\times512$ and a significantly downsized $140\times112$. Resolution downscaling was performed using bilinear interpolation implemented via OpenCV’s cv2.resize() function in Python, ensuring smooth spatial transformation while reducing dimensionality.

\begin{figure}[!t]
\centering
\includegraphics[width=\linewidth]{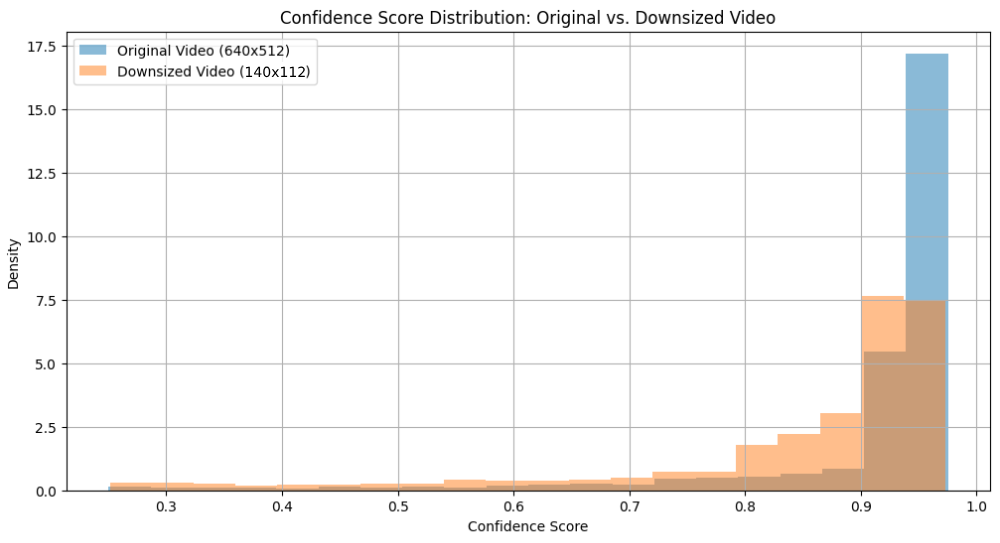}
\caption{Detection confidence over time: original vs. downsized video.}
\label{fig:confidence_plot}
\end{figure}

\begin{figure}[!t]
\centering
\includegraphics[width=\linewidth]{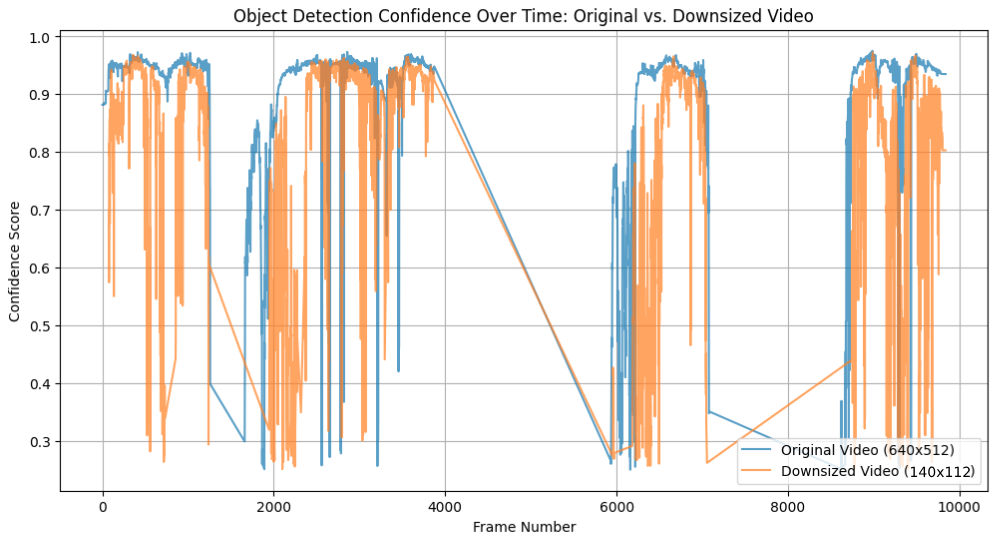}
\caption{Detection consistency over time: original vs. downsized video.}
\label{fig:consistency_plot}
\end{figure}

\begin{figure}[!t]
\centering
\includegraphics[width=\linewidth]{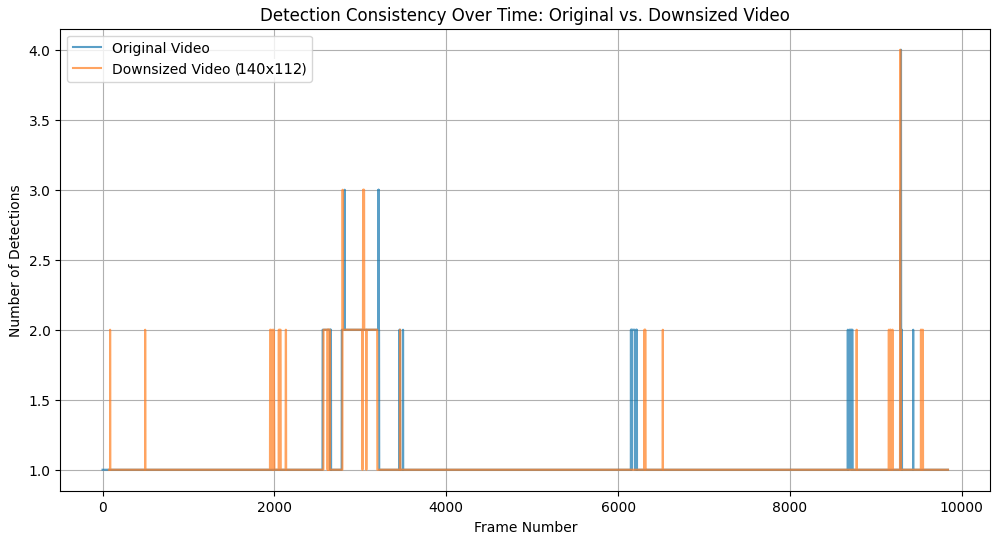}
\caption{Confidence score distribution: original vs. downsized video.}
\label{fig:confidence_distribution}
\end{figure}

The following figures illustrate how input resolution impacts detection consistency and confidence scores over time. Despite the reduced resolution, LTV-YOLO retained robust detection performance, demonstrating its potential for deployment in computationally constrained edge scenarios (Figures~\ref{fig:confidence_plot}–\ref{fig:confidence_distribution}).

The Figures indicate that LTV-YOLO maintains relatively stable detection confidence and consistency even with reduced input resolution, reinforcing its suitability for low-resource edge deployment.

\subsection{Qualitative Results and Failure Case Analysis}
\begin{figure}[t]
\centering
\includegraphics[width=\linewidth]{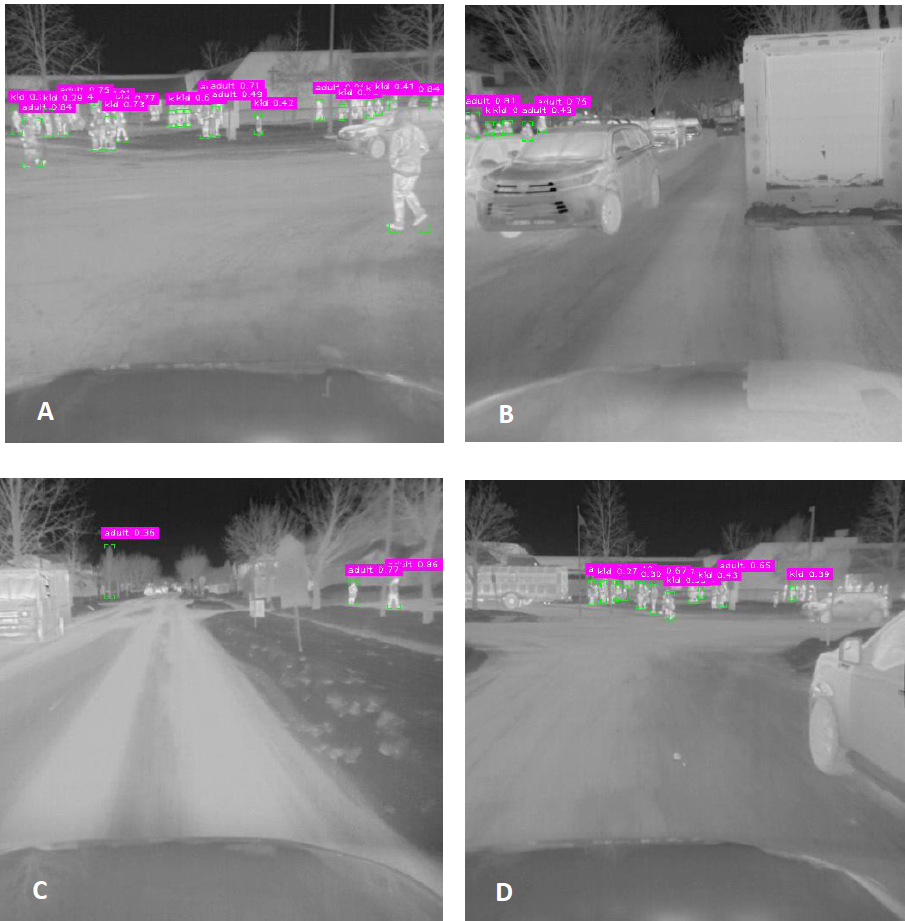}
\caption{Qualitative examples of LTV-YOLO. 
(A) Successful detection of a child pedestrian. 
(B) Successful detection of a partially occluded child. 
(C) False positive: tree detected as pedestrian. 
(D) Missed detection: distant child pedestrian.}
\label{fig:qualitative_cases}
\end{figure}

We provide qualitative evidence of model behavior in the field. 
Figure~\ref{fig:qualitative_cases} shows: (A) a successful detection of a child pedestrian; 
(B) a successful detection under partial occlusion (child behind vehicles/street furniture); 
(C) a false positive on a vertical structure near a hot background; and 
(D) a missed detection of a distant child. These examples complement the quantitative metrics.

\section{Discussion}
The experimental results demonstrate that the proposed LTV-YOLO model outperforms baseline lightweight nano-scale detectors (YOLOv5n, YOLOv8n and YOLO11n) in both accuracy and computational efficiency, particularly in detecting young VRUs. These improvements are most evident in the detection of small or partially occluded pedestrians, a crucial advancement for child safety in urban mobility systems.

A key finding in our evaluation is the superior representational capacity of our LTV-YOLO. Our model achieves 96.7\% mAP on the general benchmark of LLVIP, surpassing all the lightweight YOLO variants, including YOLOv5n (96.2\%), YOLOv8n (96.5\%) and YOLO11n (96.6\%). This superior performance can be attributed to several factors: the thermal imaging input, the use of depthwise separable convolutions for efficiency, and the integration of a Feature Pyramid Network (FPN), which enhances multiscale feature representation. Despite being the smallest model, LTV-YOLO effectively captures the complex feature representations needed for high-accuracy detection on established public benchmarks.

Our evaluation further shows that RT-DETRs achieve a marginally higher accuracy on LLVIP (97.8\%); however, it operates with approximately 20 million parameters, compared to LTV-YOLO's 1.6 million. This represents 12 times increase in model size for on 1.1\% gain in accuracy. On the other hand, LTV-YOLO delivers almost similar performance in less than 9\% of its size, making it a more practical choice for resource-constrained edge devices.

The advantages of LTV-YOLO are more visible in a custom young VRU dataset, where it achieves 91\% mAP, significantly outperforming YOLO baselines (84.5\%) and heavier RT-DETRs (81.2\%). This indicates that while the baseline models can learn standard adult pedestrian features (in LLVIP datasets), they struggle to optimize for the smaller thermal profiles of children. LTV-YOLO, a specialized design, bridges this gap, offering robust detection of young VRUs without latency penalties.

\paragraph*{Impact of Environmental Complexity}
Comparing performances across the datasets revealed a performance drop of approximately 5.7\% in mAP between the LLVIP benchmark and our custom VRU dataset. These results highlight the increased complexity inherent in the target application. While LLVIP mainly consists of high-contrast adult pedestrians in static surveillance views, our VRU dataset involves smaller subjects (children), dynamic camera motion (vehicle-mounted), and adverse weather conditions (snow). The performance gap underscores the need to collect a custom dataset, as standard benchmark datasets may not accurately reflect the detector's reliability in complex, diverse scenarios.

In summary, LTV-YOLO represents a significant step toward practical and reliable pedestrian detection systems that prioritize child safety and are deployable in edge-constrained settings. The model strikes a compelling balance between accuracy, speed, and efficiency, and sets the stage for future research in thermal-aware lightweight detection systems.

\section{Conclusion}
This paper presented LTV-YOLO, a lightweight thermal vision object detection model specifically designed to enhance the safety of young vulnerable road users (VRUs) under adverse lighting and weather conditions. Leveraging long-wave infrared (LWIR) imaging and a compact, edge-optimized architecture, LTV-YOLO addresses the limitations of conventional RGB-based detectors, particularly in low-light conditions where detecting children is often unreliable.

Our experimental results demonstrate that LTV-YOLO achieves superior accuracy, faster inference speed, and lower parameter size compared to existing lightweight detectors such as the nano variant of standard YOLO as well as RT-DETRs. The integration of depthwise separable convolutions and a Feature Pyramid Network (FPN) enabled efficient multi-scale detection, improving performance on small, partially occluded targets. Importantly, the model operates in real time on embedded devices like the Jetson AGX Orin, making it practical for deployment in smart city infrastructure, school zones, and autonomous vehicles.

While the model shows promising results, future work will focus on expanding the dataset to include rainy and foggy conditions, improving robustness through sensor fusion, and evaluating deployment at scale in live environments. The findings of this study underscore the critical role of thermal-aware, edge-capable detection systems in advancing pedestrian safety and preventing collisions involving children in complex urban settings.

\section{Acknowledgements}

The research described in this paper is jointly funded by 
the Region of Waterloo, Transport Canada and the Resilient Ground Transportation (RGT) Program of the National Research Council Canada.



\bibliographystyle{abbrvnat} 
\bibliography{references}

\section*{Author Biographies}

\newlength{\BioPhotoW}\setlength{\BioPhotoW}{1.05in}
\newlength{\BioPhotoH}\setlength{\BioPhotoH}{1.30in}
\newlength{\BioGap}\setlength{\BioGap}{0.35in}
\newlength{\BioTextW}\setlength{\BioTextW}{\dimexpr\linewidth-\BioPhotoW-\BioGap\relax}

\newcommand{\BioSep}{\par\vspace{0.85em}\noindent\rule{\linewidth}{0.3pt}\vspace{0.85em}\par}

\newcommand{\AuthorBio}[3]{%
\noindent
\begin{minipage}[t]{\BioPhotoW}
\vspace{0pt}%
\adjustbox{width=\BioPhotoW,height=\BioPhotoH,clip,center}{\includegraphics{#1}}%
\end{minipage}%
\hspace{\BioGap}%
\begin{minipage}[t]{\BioTextW}
\vspace{0pt}%
\textbf{#2}~#3
\end{minipage}\par
}

\AuthorBio{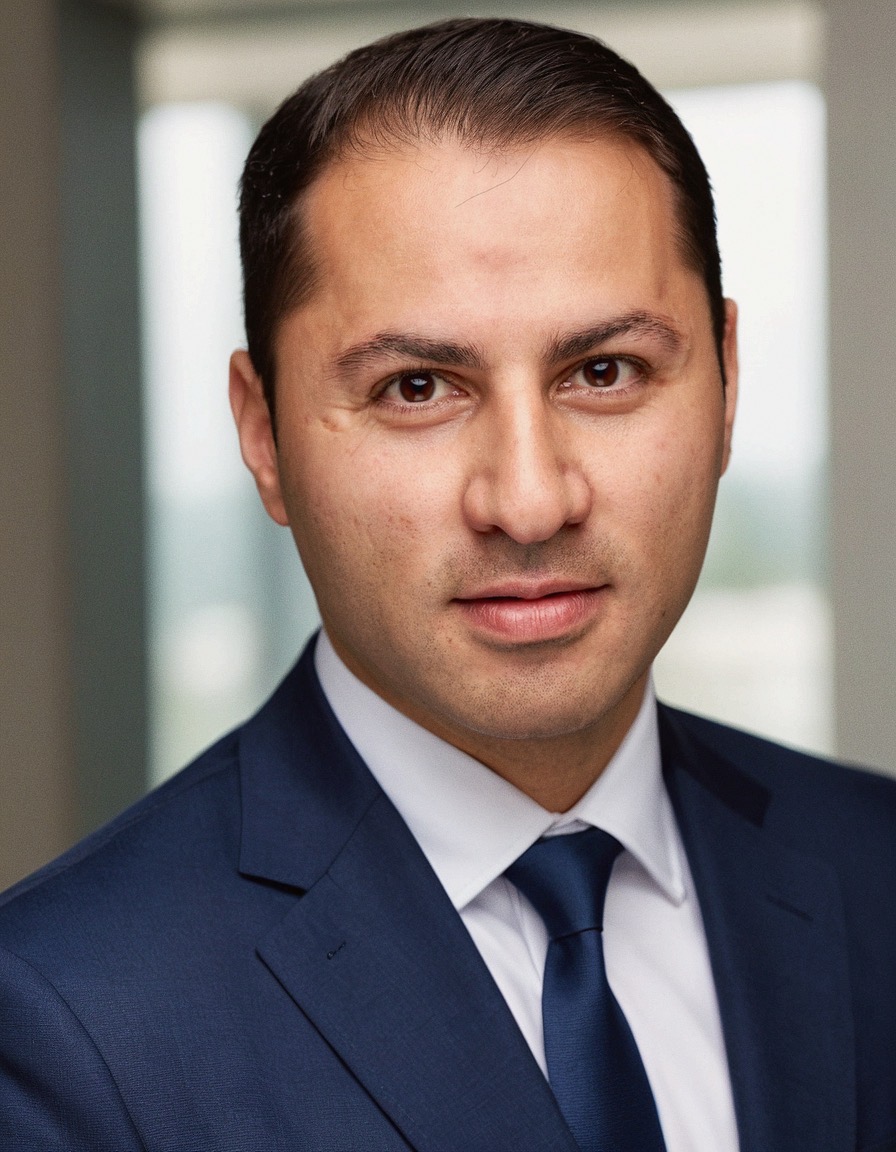}{Abdullah Jirjees}{(Member, IEEE) is currently a Research Officer in Artificial Intelligence and Computer Vision at the National Research Council Canada (NRC), where he leads applied R\&D in autonomous perception, multimodal sensor fusion, and real-time machine learning systems for connected and automated vehicles. His work focuses on embedded AI, lightweight deep learning models, LiDAR--camera fusion, and real-time perception pipelines for safety-critical transportation applications. Prior to joining NRC, he served as a Research Scientist at the Centre de Géomatique du Québec (CGQ), where he developed AI-driven geospatial analytics, UAV inspection systems, and infrastructure crack-detection tools deployed in industrial and municipal settings. He previously completed a Postdoctoral Fellowship in computer vision and deep learning at the Université du Québec à Chicoutimi (UQAC). Dr.\ Jirjees received the B.Sc.\ degree in Software Engineering in 2008 and the M.Sc.\ and Ph.D.\ degrees in Electrical and Electronic Engineering from Universiti Sains Malaysia (USM) in 2014 and 2018, respectively.}

\BioSep

\AuthorBio{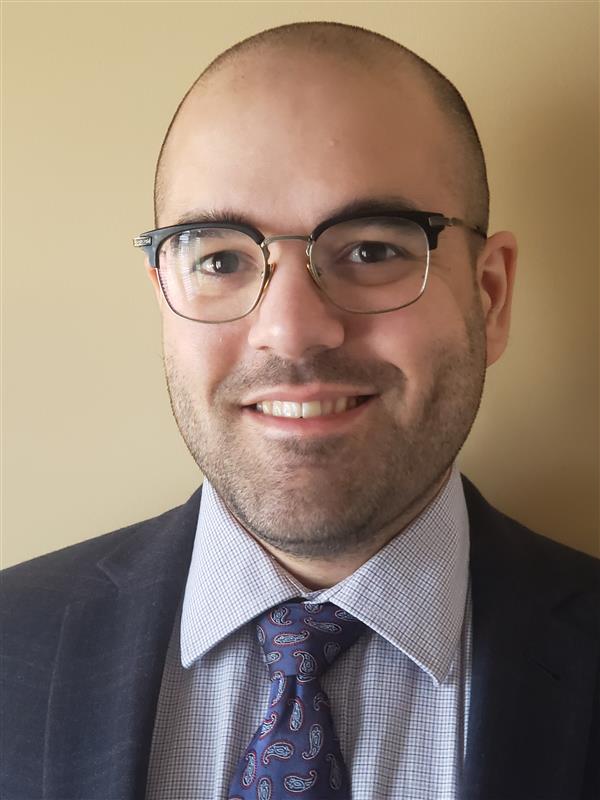}{Ryan Myers}{was born in Burlington, Ontario, Canada in 1991 and received his bachelor's degree in engineering from the University of Guelph. He is currently a thrust leader for cyber-physical manufacturing within the Advanced Manufacturing Program of the National Research Council of Canada (NRC), providing steering and technical expertise in data-driven manufacturing. His background is in mechatronic systems, data analytics, and controls engineering with over ten years of experience. As a Research Council Officer at NRC, he has served as lead expert on projects delivering process optimization and quality control for a wide range of manufacturing processes by fusing data and process expertise. Prior to NRC, he worked as a Systems and Controls Engineer, designing and deploying industrial manufacturing systems and machines, and contributing to R\&D of warehouse-focused mobile robots.}

\BioSep

\AuthorBio{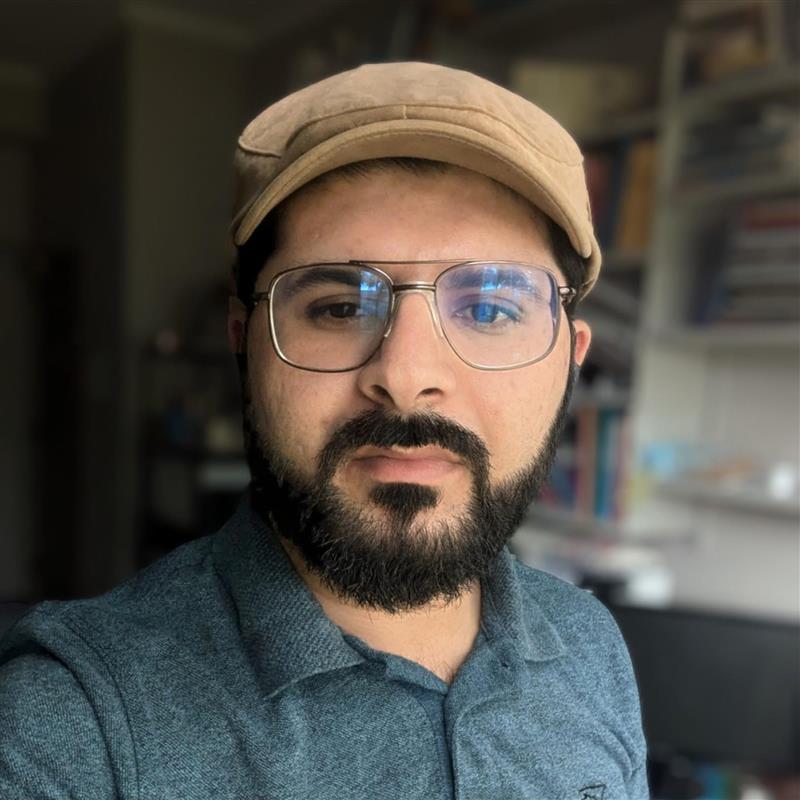}{Muhammad Haris Ikram}{received the B.Sc.\ degree in electrical engineering and the M.S.\ degree in electrical engineering from the National University of Sciences and Technology (NUST), Pakistan. He is currently pursuing the Ph.D.\ degree in civil and environmental engineering at Western University, Ontario, Canada, where his research focuses on intelligent transportation systems and data-driven safety analytics. His work integrates computer vision, machine learning, robotics, and multimodal sensing to address challenges in road safety, infrastructure monitoring, and autonomous systems. His research interests include intelligent transportation systems, machine learning, computer vision, and robotic perception.}

\BioSep

\AuthorBio{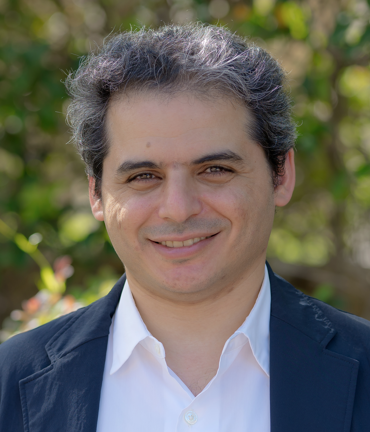}{Mohamed H. Zaki}{(Member, IEEE) is an Associate Professor in the Civil and Environmental Engineering Department at Western University, Ontario, Canada. Previously, he was an Assistant Professor in the Civil, Environmental \& Construction Engineering Department at the University of Central Florida, Orlando, Florida. Dr.\ Zaki is a member of IEEE and has served on the Transportation Research Board (TRB) AED50 Committee on Artificial Intelligence and Advanced Computing Applications.}

\end{document}